\documentclass[conference]{IEEEtran}

\usepackage{amsmath,amssymb,amsfonts}
\usepackage{algorithmic}
\usepackage{graphicx}

\usepackage{url,microtype,tabularx}
\usepackage{microtype}                 
\PassOptionsToPackage{warn}{textcomp}  
\usepackage{textcomp}                  
\usepackage{mathptmx}                  
\usepackage{times}                     
\newcommand{\citep}{\cite}
\usepackage{cite}                      
\usepackage{booktabs}
\usepackage{pgfplots}
\usepackage{multirow}
\usepackage[colorlinks=true,linkcolor=black,anchorcolor=black,citecolor=black,filecolor=black,menucolor=black,runcolor=black,urlcolor=black]{hyperref}
\usepackage{floatrow}

\def\BibTeX{{\rm B\kern-.05em{\sc i\kern-.025em b}\kern-.08em
    T\kern-.1667em\lower.7ex\hbox{E}\kern-.125emX}}
\usepackage{orcidlink}
\usepackage{todonotes}
\usepackage[commentmarkup=uwave,final]{changes}
\definechangesauthor[name=Marc, color=orange]{ml}
\definechangesauthor[name=Andreas, color=red]{ah}
\definechangesauthor[name=Chris, color=violet]{cs}

\title{Comparison of Data Encodings and Machine Learning Architectures for User Identification on Arbitrary Motion Sequences}

\author{

\IEEEauthorblockN{Christian Schell \orcidlink{0000-0002-0022-0711}}
\IEEEauthorblockA{
    \textit{University of Würzburg, Germany}
}
\and
\IEEEauthorblockN{Andreas Hotho \orcidlink{0000-0002-0483-5772}}
\IEEEauthorblockA{
    \textit{University of Würzburg, Germany}
}
\and
\IEEEauthorblockN{Marc Erich Latoschik \orcidlink{0000-0002-9340-9600}}
\IEEEauthorblockA{
    \textit{University of Würzburg, Germany}
}

}

\begin{document}

\maketitle

\begin{abstract}
Reliable and robust user identification and authentication are important and often necessary requirements for many digital services.
It becomes paramount in social virtual reality (VR) to ensure trust, specifically in digital encounters with lifelike realistic-looking avatars as faithful replications of real persons.
Recent research has shown that the movements of users in extended reality (XR) systems carry user-specific information and can thus be used to verify their identities.
This article compares three different potential encodings of the motion data from head and hands (scene-relative, body-relative, and body-relative velocities), and the performances of five different machine learning architectures (random forest, multi-layer perceptron, fully recurrent neural network, long-short term memory, gated recurrent unit).
We use the publicly available dataset ``Talking with Hands'' and publish all code to allow reproducibility and to provide baselines for future work.
After hyperparameter optimization, the combination of a long-short term memory architecture and body-relative data outperformed competing combinations: the model correctly identifies any of the 34 subjects with an accuracy of 100\% within 150 seconds.
Altogether, our approach provides an effective foundation for behaviometric-based identification and authentication to guide researchers and practitioners.
Data and code are published under \url{https://go.uniwue.de/58w1r}.
\end{abstract}

\section{Introduction}


The past decade has seen a significant boost in new developments and technologies for Virtual, Augmented, and Mixed Reality (VR, AR, MR: XR for short). A large variety of consumer-grade hardware devices and software solutions are now affordable, opening up interesting and promising out-of-lab use-cases. Here, many of these use cases benefit from social XR \citep{Latoschik2019NotReality}. Social XR provides avatar-embodied encounters of users in an unlimited number of different virtual environments. 
Most prominently, social XR promises mediated human-to-human interaction over distances while supporting the richness of verbal and nonverbal interaction based on users' avatars. Altogether, these benefits render social XRs highly attractive, not least because of the Covid-19 pandemic and the necessary reduction or even avoidance of physical contacts.

Recently, technological advances in creating lifelike virtual humans have made considerable progress. For example, the MetaHuman project \citep{metahuman} provides a variety of high-fidelity digital humans indistinguishable from real persons.
In addition, photogrammetry-based methods \citep{Achenbach2017FastHumans,realistic_humans_smartphone,latoschik2017effect} provide individualized near-photorealistic avatars that faithfully replicate the real physical visual appearance of users.
Driven by these developments, Lin and Latoschik \cite{lin2022digital} formulate pressing questions in their recent state-of-the-art review of identity and privacy in social VR:
how can we trust the digital alter egos of others and verify their real identity?
A general question that becomes more and more important during times of fake news and deep fake approaches.

An interesting new way to check users' identities is to analyze their tracking data.
This continuous stream of data is inherently available to any XR system, since it is required for immersion and user interaction.
Recent work, listed in \autoref{tab:related_work_comparison}, has already demonstrated that movement data of XR users carries identifying information.
For example, Miller M. et al. \cite{Miller2020a} could identify 511 VR users watching 360° videos with over 95\% accuracy.
However, as Stephenson et al. \cite{Stephenson2022SoKReality}, as well as Lin and Latoschik \cite{lin2022digital}, point out in their meta-reviews, this field of research is still in its early stages, and practical real-world solutions have yet to be developed.
In this article, we focus on two fundamental aspects of XR user identification to guide future work in their decisions:

\begin{enumerate}
    \item \emph{Feature pre-processing}: we test three different data encodings of the movement data (i.e., scene-relative, body-relative, and body-relative velocity).
    
    \item \emph{Machine learning architectures}: we compare five different prominent architectures for the identification task, i.e., random forest, multi-layer perceptron, and three types of deep recurrent neural networks (fully recurrent neural network, long short-term memory, and gated recurring unit).
\end{enumerate}

For this we use the head and hand tracking data from the publicly available ``Talking With Hands'' dataset \citep{Lee2019TalkingSynthesis}.
After an extensive hyperparameter search, deep learning approaches outperform the random forest and multi-layer perceptron and achieve 100\% identification accuracy within 150 seconds on test sequences on body-relative as well as body-relative velocity data.
The results also show that no machine learning model generalizes well using scene-relative data, which renders this data encoding problematic for identification purposes.
Altogether, our contribution provides an effective foundation for behaviometric-based identity checks based on deep learning of arbitrary motion data sequences for XR use-cases.
\section{Related Work}
We have summarized relevant related work in \autoref{tab:related_work_comparison} to guide the following discussion. Several works started to exploit motion data as generated by typical VR setups \cite{Miller2020a,Mustafa2018,Kupin2019,Pfeuffer2019, Ajit2019,Mathis2020,Miller2020,Miller2021,Olade2020,Liebers2020,Moore2021PersonalSessions}.
Such setups often provide at least a basic three-point tracking: head tracking is essential to enable visual immersion, while hand tracking is required for user interaction.
Additionally, motion data from other sources, like the IMU sensors in Google Glasses, can be used for identification purposes as well \cite{Rogers2015, Li2016,Shen2019}.

In this paper we focus on identification. Note, that identification and authentication are not the same. Like Miller M. et al. \cite{Miller2020a} we also follow the definition by Rogers et al. \cite{Rogers2015}:
authentication focuses on independently identifying a user from any other potential user and is usually most concerned with avoiding false-positive classifications.
Identification is the task of identifying a user from a known group of users and equally rates false-positive and false-negative classifications.

Previous works consider a wide range of tasks for their use-cases (see column ``task'' in \autoref{tab:related_work_comparison}). This results in recordings from actions that range from very specific to non-specific.
Sequences of \textit{specific} actions involve the user performing an instructed movement, like throwing a ball \cite{Liebers2021,Ajit2019,Miller2019RealtimeEnvironments}, watching rapidly changing images \cite{Rogers2015}, or nodding to music \cite{Li2016}.
More general use cases, like troubleshooting a virtual robot \cite{Moore2021PersonalSessions} or watching 360° videos \cite{Miller2020a}, lead to sequences that involve the user performing non-specific actions, so it is not well defined what the user is doing within a given time span.
Such \textit{arbitrary} movements are arguably a more challenging basis for user identification, since models must learn identifying patterns within different user movements, regardless of the actual action.
Yet, models that can generally identify users without the requirement for specific movements are also more interesting, since these are more flexible and could be used to establish identification services running in the background, monitoring the continuously produced data streams.

\begingroup
\setlength{\tabcolsep}{5pt} 
\renewcommand{\arraystretch}{1.25} 

\begin{table*}[t]
\caption{\label{tab:related_work_comparison} Relevant work targeting identification or authentication based on machine learning of movement data; $N$ describes the number of individual user; SR = scene relative, SRV = scene relative velocity, BR = body relative; HMD = Head Mounted Display.}

\resizebox{\textwidth}{!}{%
\begin{tabular}{l|>{\raggedright\arraybackslash}p{2.5cm}|>{\raggedright\arraybackslash}p{4cm}|>{\raggedright\arraybackslash}p{3cm}|l|>{\raggedright\arraybackslash}p{2cm}}

\rule{0pt}{2ex}
\textbf{authors}                & \textbf{classifier}                          & \textbf{task}                                                                     & \textbf{data encoding}                                                                           & \textbf{dataset}             & \textbf{device}                                  \\[1ex]

\midrule

Rogers et al. (2015) \cite{Rogers2015} & random forest                       & ident.: watching rapidly changing images            & acc. \& orient. of head + eye blinking                                         & N=20; unpubl.  & Google Glass                             \\

Li et al. (2016) \cite{Li2016} & distance-based                      & auth.: nodding to music                                    & acceleration of head                                                                     & N=95; unpubl.  & Google Glass                             \\

Mustafa et al. (2018) \cite{Mustafa2018} & logistic regression, svm            & auth.: walking                                & acc. \& orient. of HMD                                                        & N=23; unpubl.  & Google VR                                \\

Kupin et al. (2019) \cite{Kupin2019} & nearest neighbor, distance-based    & auth.: ball throwing                                                   & SR of right controller                                                                   & N=14; unpubl.  & HTC Vive                                 \\

Pfeuffer et al. (2019) \cite{Pfeuffer2019} & random forest, svm                  & ident.: point, grab, walk, type                                          & SR, SRV of HMD \& controllers                          & N=22; unpubl.  & HTC Vive                                 \\

Shen et al. (2019) \cite{Shen2019} & distance-based                      & auth.: walking a few steps                                               & acc. \& orient. of head                                                        & N=20; unpubl.  & Google Glass                             \\

Ajit et al. (2019) \cite{Ajit2019} & nearest neighbor, distance-based    & auth.: ball throwing                                                   & SR of HMD \& contr.                                                               & N=33; unpubl.  & HTC Vive                                 \\

Mathis et al. (2020) \cite{Mathis2020} & fully conv. network         & auth.: interaction with a cube                                        & SR of controllers                                                                        & N=23; unpubl.  & HTC Vive                                 \\

Miller M. et al (2020) \cite{Miller2020a} & random forest, knn, gbm    & ident.: watching 360° videos and answering questionnaire                 & SR of controllers and head                                                               & N=511; unpubl.  & HTC Vive                                 \\

Miller R. et al. (2020) \cite{Miller2020} & distance-based                      & auth.: ball throwing                                                     & SR, SRV, trigger position of controllers       & N=41 publ.  & HTC Vive \& Vive Cosmos, Oculus Quest  \\

Olade et al. (2020) \cite{Olade2020} & nearest neighbor                    & auth. \& ident.: grab, rotate, drop balls and cubes                      & SR of HMD \& contr. + eye gaze                                                    & N=15; unpubl.*  & HTC Vive                                 \\

Miller R. et al. (2021) \citep{Miller2021} & siamese nn & like \cite{Miller2020} & SR of HMD \& contr. & like \cite{Miller2020}     & like \cite{Miller2020} \\

Liebers et al. (2021) \cite{Liebers2021} & LSTM, MLP                               & ident.: bowling, archery                                                  & BR of HMD and controllers                                                                & N=16; publ.     & Oculus Quest                                 \\

Moore et al. (2021) \cite{Moore2021PersonalSessions} & random forest, knn, gbm                           & ident.: robot troubleshooting  in VR      &  SR and SRV of controllers and HMD                                                                & N=60; unpubl.     & HTC Vive                            \\

Miller R. et al. (2022) \citep{Miller2022CombiningBiometrics} & siamese nn & like \cite{Miller2020} & SR of HMD \& contr. & like \cite{Miller2020} & like \cite{Miller2020} \\

Miller R. et al. (2022) \cite{Miller2022TemporalBiometrics} & siamese nn & like \cite{Miller2020} & SR of HMD \& contr. & like \cite{Miller2020} \& \cite{Ajit2019} & like \cite{Miller2020} \\
\midrule
this paper & random forest, MLP, FRNN, LSTM, GRU             & ident.: conversation                                                     & SR, BR, BRV of head and hands                                                            & N=34; publ.     &  3-point tracking from full body mocap \\
\end{tabular}}
\footnotesize *authors indicated a publication of their data, but there has been none so far
\end{table*}
\endgroup

The majority of prior work was using groups of 14 to 95 individuals to develop identification methods. The recent work by Miller et al. \cite{Miller2020a} is the first to provide convincing evidence that identification can be performed even on a large group of 511 individuals. The authors have trained a random forest model on raw motion data referenced in the coordinate system of the 3D scene (scene-relative). The model correctly classified 95\% of the withheld test sequences. The same methods were evaluated by Moore et al. \cite{Moore2021PersonalSessions} on a different dataset and the authors agree that arbitrary movement data can be identifying.

Altogether, we identified three common shortcomings of previous work and cover them in this paper: first, we follow Miller M. et al. \cite{Miller2020a} and Liebers et al. \cite{Liebers2020} in arguing that proper pre-processing of motion data features is important to help classification models learn user-specific instead of session-specific patterns. This might be a severe potential weakness of the --- otherwise convincing --- results of Miller M. et al \cite{Miller2020a}, yet little work has been done to systematically explore this specific matter. Second, there are only few works exploring deep learning methods and their beneficial implicit feature processing and selection  \cite{Liebers2020,Mathis2020,Miller2022TemporalBiometrics}. However, their approaches lack a convincing hyperparameter search and comparison to basic classifiers used by other authors, such as random forest. Consequently, it remains unclear how increased complexity and computational cost of deep learning architectures can be weighed against potentially better performances. Lastly, we think it is problematic that except of Liebers et al. \cite{Liebers2021} and Miller R. et al. \cite{Miller2022TemporalBiometrics} none of the previous work uses publicly available data which makes it impossible to compare different approaches. In addition, to our knowledge, only Miller R. et al. \cite{Miller2022TemporalBiometrics} have published their code for data processing and training, which compromises verifiability and reproducibility of all other previous publications.

Against this backdrop, we investigate the effects of three different motion data encodings (i.e., scene-relative data, body-relative data and body-relative velocity data) combined with five prominent machine learning architectures (random forest, multi-layer perceptron and three different types of recurrent neural networks). We propose a structured approach to train capable deep learning models for behaviometric-based user identification. To allow replication of our work we use the publicly available ``Talking With Hands'' dataset from Lee et al. \cite{Lee2019TalkingSynthesis} and publish the code we have developed for the data encoding and machine learning steps under \url{https://go.uniwue.de/58w1r}.
\section{Methodology}

\subsection{Posture and Movement Data Encoding}

3D engines, such as Unity or the Unreal Engine, provide access to a steady stream of positions and rotations of tracked devices or joints from the 3D motion sensors. This data is always specified with respect to a coordinate frame of reference. In general, such reference frames are free to choose, and they can be mapped from one to the other using appropriate 3D-transformations. Usually, the engine's scene root serves as the default reference. We therefore refer to this type of data as \emph{scene-relative} (SR) data. This choice of the frame of reference corresponds to the approaches of Miller M. et al. \cite{Miller2020a}, Ajit et al. \cite{Ajit2019}, Moore et al. \cite{Moore2021PersonalSessions} and Mathis et al. \cite{Mathis2020}. SR data frames consist of 21 features: (pos-x, pos-y, pos-z, rot-x, rot-y, rot-z, rot-w) $\times$  (head, wrist-left, wrist-right) all given with respect to the scene root as frame of reference. Note, that rotations are represented as quaternions.

However, SR-encoded data does not only incorporate user-specific characteristics directly correlating to the user's identity.
For example, the user's absolute position and orientation in the VR scene are, at least partially, arbitrary: 
they can change due to navigation (e.g., teleportation), new frames of reference (e.g., level changes) or new calibrations between usages --- all reasons the values of tracking data change without conveying anything user-characteristic.
Unfortunately, it is easy for a model to optimize towards these characteristics that are not user-specific.
To remove any information about the whereabouts of the user, we transform the coordinate system from scene-relative to the \emph{body-relative} (BR) frame of reference: 
position and orientation of the wrists are defined with respect to the head, which results in the wrists being positioned and orientated independently of the user's original scene position and orientation. The head's positional features and the rotation around the up axis become obsolete and are therefore removed, which only leaves one quaternion encoding the head's rotation around the horizontal axes. This has the effect that the data yields the same values for the same movement (e.g., waving), even if the user changes position or orientation within the scene in between.
BR data consists of 18 features: (pos-x, pos-y, pos-z, rot-x, rot-y, rot-z, rot-w) $\times$  (wrist-left, wrist-right) + (rot-x, rot-y, rot-z, rot-w) $\times$ (head), all given with respect to the user's head as frame of reference. 


Even though BR data contains less information than SR data, it may still contain information that is irrelevant or even misleading for identifying the originator. For example, the data encodes the exact distance and angle of the hands to the head at each point in time: if a user repeats the same motion, the joints might follow the same trajectory, but start and end at different positions w.r.t. the body. In theory, machine learning models can learn to deal with such variances, but require an appropriate amount of training data. It might therefore be beneficial to remove even more information from the data by computing \emph{body-relative velocity} (BRV) data: for positions we subtract the value of each feature with the value of the corresponding value of the previous frame, for rotations we rotate each quaternion by the corresponding inverted rotation of the previous frame.
This way the data only contains a signal, when the user is actually moving.
The BRV data also consists of 18 features, just like the BR data.

We hypothesize that the BRV encoding can help the classification models learning user-specific characteristics by removing unneeded information (i.e., noise).
Note, that Moore et al. \cite{Moore2021PersonalSessions} seem to contradict this consideration by demonstrating that their models perform considerably worse when trained on velocity data.
However, the authors computed the velocities based on scene-relative data (just like Pfeuffer et al. \cite{Pfeuffer2019} and Miller et al. \cite{Miller2020}), so the resulting features still encode session specific characteristics, since the orientation of the user remains captured within the data.

It might be worthwhile to further investigate acceleration or even jerk data as well.
However, adding further data encodings would increase the hyperparameter search space later described.
Hence, to keep the scope of this work within reasonable limits, we focus on the three presented data encodings.

\subsection{Machine Learning Architectures}

Previous works investigate a wide range of machine learning architectures (see column ``classifier'' in \autoref{tab:related_work_comparison}).
For this article we aimed to select a broad spectrum of architectures:

\begin{enumerate}
    \item Random Forest (RF)
    \item Multi-layer Perceptron (MLP)
    \item Fully Recurrent Neural Network (FRNN)
    \item Long-Short Term Memory (LSTM)
    \item Gated Recurrent Unit (GRU)
\end{enumerate}

Random forests can be seen as basic machine learning models that provide few hyperparameters and are relatively easy to train.
MLPs are the most basic neural networks that can be layered and structured in arbitrary ways to form powerful pattern extractors.
However, MLPs cannot work directly with time series data, and thus need the same preprocessing as random forests.
Recurrent neural networks, such as the FRNN, LSTM and GRU, are deep neural networks that can work directly with the time series data and might therefore be able to extract more complex patterns.

Each architecture provides several parameters that configure either the architecture itself or the training process.
These so called hyperparameters have profound impact on the capabilities of the resulting models and are individual to each problem and dataset. We provide an overview of the selected hyperparameters and the search space we explore in \autoref{tab:hyperparameters}.

\begin{table}
	\caption{Relevant parameters selected for the hyperparameter search - The architectural parameters explored during hyperparameter search stage 1.}
    \begin{tabularx}{\linewidth}{p{2.75cm}ll}
                         & \textbf{hyperparameter} & \textbf{search space}  \\ 
\midrule
\textbf{random forest}   & n\_estimators           & 50 - 1,000 \\
                         & min\_samples\_leaf      & 1 - 1,000 \\ 
\midrule
\textbf{MLP}  			 & number of layers        & 1 - 6             \\
                         & size per layer          & 10 - 300 neurons  \\
                         & learning rate           & 0.00001 - 0.01 \\
\midrule
\textbf{FRNN, LSTM, GRU} & hidden size             & 20 - 200 neurons \\
                         & number of layers        & 1 - 8                  \\
                         & dropout                 & 0 - 0.6 \\
                         & learning rate           & 0.0001 - 0.01
\end{tabularx}
    \label{tab:hyperparameters}
\end{table}
\begin{table}
	\caption{Relevant parameters selected for the hyperparameter search - The data parameters explored during hyperparameter search stage 2; the original framerate is 90 FPS.}
    \begin{tabularx}{\linewidth}{p{2.75cm}ll}
                 & \textbf{hyperparameter} & \textbf{search space}  \\
\midrule
\textbf{bins}    & frames per bin          & 10 - 1,350 frames \\
\midrule
\textbf{windows} & frames per second       & 10, 30, 60, 90 fps \\
                 & window size             & 10, 100, 300 frames
\end{tabularx}
	\label{tab:data_hyperparameters} 
\end{table}

\subsection{Data Sampling}

We divide the recordings of each participant into subsequences of a given length (e.g., 10 seconds).
Depending on the machine learning architecture these subsequences have to be transformed into a format the algorithms can work with.
In this section we describe two sampling approaches, one for RF and MLP, the other for the RNNs.
Additionally, we define hyperparameters for each approach that determine how long the subsequences are (\autoref{tab:data_hyperparameters}).
In both cases, features are scaled to zero mean and a variance of one, based on the training data set, which helps the machine learning training process.

\subsubsection{Binned Samples for Random Forest and MLP}
To transform the time series data into one-dimensional samples, we adopt the methodology of Miller M. et al \cite{Miller2020a}: we bin the frames of each recording into chunks of a fixed time period and then reduce each chunk to a selection of five statistics per feature: minimum, maximum, mean, median and standard deviation. Consequently, a binned sample either has 105 features for SR data or 90 features for BR and BRV data.

\subsubsection{Windowed Samples for Recurrent Neural Networks}
The RNNs can work directly with time series data. The only pre-processing required is to organize the frames into windowed samples. For windowed samples we investigate two hyperparameters. Firstly, there is the \textit{window size}. While in theory RNNs would work with arbitrarily long sequences, in practice there has to be a fixed window of frames to make training technically feasible. Secondly, there is the number of \textit{frames per second}. The recordings of the Talking With Hands dataset are sampled at 90 frames per second (fps). To measure the effects of lower temporal resolutions we also downsample the data to lower frequencies. Consequently, the time period of a windowed sample depends on both hyperparameters. For example, a window size of 30 frames and a frequency of 10 fps results in a period of $\frac{30 \text{frames}}{10 \text{fps}} = 3$ seconds per sample.

\subsection{Prediction}
\label{sec:prediction}
Once a model is trained, it can be used for inference.
The models receive either a binned or a windowed sample to predict one of the 34 subjects.
In our evaluation we also consider a majority voting approach:
for a given sequence we take the predictions the model produced for each sample and select the most commonly predicted class as final prediction.
We repeat this for all possible sequences within each test take and report the averaged results.
For this evaluation we consider sequence lengths ranging from the number of frames required for one sample, up to 27.000 frames (i.e., 5 minutes).
Note, that we expect any model to get 100\% accurate eventually, given that the model predicts each class better than random --- the sequences for the majority voting just have to be long enough.

\section{Experimental Setup}

In our experiment we consider a use case where we observe users talking to each other, a likely scenario in the context of social VR.
Since there are no specific actions the users are instructed to perform, the sequences of tracking data we perform the identification on are arbitrary.
This means that for any given subsequence a model sees, it is not defined what action the user is doing at that moment.
In the following sections we describe the used dataset, the data preparation and details about training and evaluation.

\subsection{Dataset}
We use the ``Talking with Hands'' dataset from \cite{Lee2019TalkingSynthesis}, since it is to our knowledge the largest publicly available dataset containing human motion capture data. It contains approximately 20 hours of footage recorded with a full body motion capture system from 37 individual subjects recorded over 32 sessions. In each session, two subjects have been recorded performing several conversational tasks. A session consists of multiple takes, each containing either recordings of conversations of the two subjects, or of system calibration tasks performed by one subject. 

The recordings contain tracking data of the full body. To emulate three-point tracking we select three bones from the tracked skeleton that are closest to the sensors of a typical XR setup: the bone ``b\_head'' to represents the HMD and the bones ``b\_l\_wrist\_twist'' and ``b\_r\_wrist\_twist'' the respective controllers.

The authors of the dataset used a standardized skeleton to project each subject's movements on to, which removes any explicit information about body proportions.
This is different to typical XR setups using HMDs and controllers, where the tracking data gets projected directly onto virtual heads and hands and therefore reflects, for example, a person's height. 
Consequently, in the data used in this paper there is no explicit information about individual body heights or arm lengths.
However, body proportions incorporate significant information attributable to the identity of users, such as the body height or the length of extremities.
Hence, it is safe to assume that the developed solutions could even be improved when this standardization by an enforced retargeting would not be applied.

We also considered evaluating our approaches with the data from Liebers et al. \cite{Liebers2021} and Miller R. et al. \cite{Miller2022TemporalBiometrics}. However, there is less than 4 minutes of recording per user and feature set in each of the datasets. This made it difficult for us to find a legitimate way to split the data into training, validation and test sets for our methods to allow a meaningful comparison.

We have also contacted the authors of Miller M. et al. \cite{Miller2020a} and Moore et al. \cite{Moore2021PersonalSessions} for access to their datasets. Unfortunately, due to missing user agreements neither authors were able to share their data.

\subsection{Data Filtering and Splitting}
Takes used for training and evaluating models should contain only recordings of actual conversations and exclude any calibration takes.
Since the takes are not labeled, we have conversed with the authors of the dataset and followed their recommendation: 
only takes are selected that 1) are from scenes with two subjects, 2) have a minimum length of five minutes and 3) show a minimum indication of movement. 
The last point is necessary, because we encountered corrupted takes containing frozen armatures that do not move at all. 
The authors have been made aware of this issue and may update the dataset in the future.

We divide the dataset into three subsets: the \emph{training set} is used for training the models.
Due to overfitting, models can perform well on training data, but may not generalize well to unseen data.
We therefore use a \emph{validation set} to evaluate the performance during training and to compare performances of different configurations during hyperparameter search. 
Finally, we use the \emph{test set} to report and discuss the results of the final models.

Consequently, we require a minimum of three takes per subject, one for each subset. 
For the split, we sort the takes of each subject by their length and use the shortest for testing and the second shortest for validation. 
The remaining takes of a subject are used for training. 
This way there is one take per subject for each validation and test set with a minimum length of 5 minutes.
Five subjects appear in more than one session, for these we select the longest session and ignore the remaining sessions.
From the 37 subjects, 34 remain after filtering. 
Overall, the total length of footage used per subject ranged from 22 minutes to 70 minutes and in one case even exceeded 90 minutes, as the authors of the dataset did not impose any constraints on the number or duration of takes for each subject.

%

\subsection{Implementation and Training}

All models are trained on a computing cluster of the computer science department of our university. We use PyTorch Lightning \cite{Falcon_PyTorch_Lightning_2019} to implement MLP, FRNN, LSTM and GRU in Python. For optimization we use Adam with the categorical cross-entropy as loss criterion. We train each model for a maximum of 300 epochs, as preliminary runs did hardly improve beyond that. Additionally, we stop training early when the train loss drastically deteriorates after an initial grace period. During training we save a snapshot of each model at its validation highpoint for later evaluation since performance can decline towards the end. Each PyTorch training job runs with 8 CPU units, one GPU (either NVIDIA GTX 1080Ti, RTX 2070 Ti or RTX 2080Ti) and 20 Gb RAM. For random forest we use the implementation of the scikit-learn library and train it with the default settings of scikit-learn's fitting method. The implementation requires substantial RAM and does not benefit from GPUs, so training jobs run with 8 CPU units and 30 Gb RAM.

\subsection{Evaluation Metric}

During initial trial runs we used the \emph{mean accuracy} averaged over all subjects to pick winner models.
This is in line with previous work that is concerned with identification tasks \citep{Pfeuffer2019,Olade2020}: 
the accuracy of one subject $s$ is given by the ratio of true positive classifications $TP$ to the number of all subject samples $n_s$: $Acc_{s}  = \frac{TP_s}{n_s}$.
Then, we report the macro averaged accuracy, which accounts for eventual class imbalances: $S$: $MeanAcc = \frac{1}{S} \sum_{s=1}^{S} Acc_s$.

However, we noticed that this metric is not suitable to select reliable models, since classes with high accuracy compensate for classes that the model produces mainly false negatives for.
We therefore select the \emph{minimum accuracy} as target metric for the hyperparameter search.
The minimum accuracy is the smallest of all $Acc_s$: $MinAcc = \min(Acc_1, Acc_2, ..., Acc_{S})$. 
This way we prefer models that work decently for all subjects over models that detect most, but not all, subjects.

\subsection{Hyperparameter Search}

We perform a hyperparameter search for each of the 15 combinations of data encodings (SR, BR and BRV) and architectures (RF, MLP, FRNN, LSTM, GRU). We use the open source hyperparameter optimization framework \emph{Optuna} \cite{Akiba2019} to generate hyperparameter configurations from the search space. The hyperparameter configuration of any combination consists of 1) settings for the \emph{architectural hyperparameters} (\autoref{tab:hyperparameters}) as well as 2) the \emph{data hyperparameters} (\autoref{tab:data_hyperparameters}). We separate the hyperparameter search for these two categories into two stages to keep the search space manageable.

In the first stage, we use Optuna's default sampler to propose configurations from the search space (as defined in \autoref{tab:hyperparameters}) for training. During this stage, the data hyperparameters remain fixed to a duration of one second (following \cite{Miller2020a}): binned samples are set to 90 frames per bin, windowed samples are set to a window size of 30 frames and a resolution of 30 frames per second.

In the second stage, we select for each combination the configuration that scored the best weighted accuracy on the validation data and continue with that configuration for the data hyperparameter search.
In this stage we perform one training for each combination of the data hyperparameter search space.

\section{Results}
\newcommand{\resultDirectory}{2021-08-21}

\label{sec:architectural_hs}

\begin{table}
\footnotesize

\caption{Final configurations after architecture and data hyperparameter search; abbreviations: `ne' is n\_estimators; `msl.' is mean\_samples\_leaf; `lr' is learning rate; `l. $\times$ ls.' is layers and layer size; `do' is dropout.}
\label{tab:configurations}
\begin{tabular}{ll|rrr|rrl}
\toprule
 &  & \multicolumn{3}{l|}{\textbf{model configuration}} & \multicolumn{2}{l}{\textbf{sample conf.}} \\
 
\midrule
\midrule
 &  & \textsc{ne} & \textsc{msl} & & frames & fps \\
\midrule
\multirow[t]{3}{*}{\textbf{RF}} & SR  & 85  & 78 & & 90  & - \\
                       & BR  & 180 & 930 & & 700 & - \\
                       & BRV & 78  & 2   & &  700 & - \\
 
\midrule
\midrule
 &  & \textsc{lr} & \textsc{l.} $\times$ \textsc{ls.} & \textsc{do} &  &  &  \\
\midrule
\multirow[t]{3}{*}{\textbf{MLP}} & SR   & 0.01 & 1 x 204 & - & 120 & - \\
                        & BR   & 0.05 & 3 x 18  & - & 900 & - \\
                        & BRV  & 0.01 & 6 x 83  & - & 1350 & - \\

\midrule
\multirow[t]{3}{*}{\textbf{FRNN}} & SR & 0.001 & 5 x 180 & 0.55 & 30 & 30 \\
                         & BR & 0.001 & 4 x 20 & 0.34 & 30 & 30 \\
                         & BRV & 0.0005 & 1 x 200 & 0.00 & 100 & 10 \\
 
\midrule
\multirow[t]{3}{*}{\textbf{LSTM}} & SR & 0.01 & 4 x 160 & 0.05 & 30 & 30 \\
                         & BR & 0.003 & 3 x 20 & 0.26 & 300 & 60 \\
                         & BRV & 0.0005 & 5 x 160 & 0.01 & 300 & 30 \\
 
\midrule
\multirow[t]{3}{*}{\textbf{GRU}}  & SR & 0.007 & 2 x 140 & 0.08 & 30 & 30 \\
                         & BR & 0.006 & 5 x 60 & 0.50 & 100 & 10 \\
                         & BRV & 0.002 & 3 x 160 & 0.05 & 300 & 30 \\
\bottomrule
\end{tabular}
\end{table}

\begin{figure}[h]
	\centering
	\includegraphics[width=0.98\linewidth]{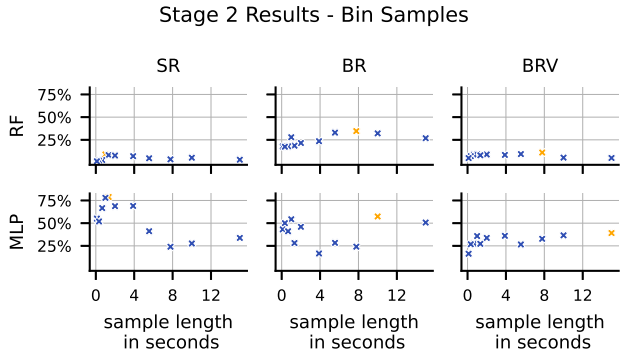}
	\caption{Results of the data hyperparameter search for binned samples (see \autoref{tab:data_hyperparameters}), indicating the achieved minimum accuracy of each configuration on the validation data in percent.}
	\label{fig:results_data_hps_binned}

\end{figure}

\begin{figure}[h]
		\centering
		\includegraphics[width=0.98\linewidth]{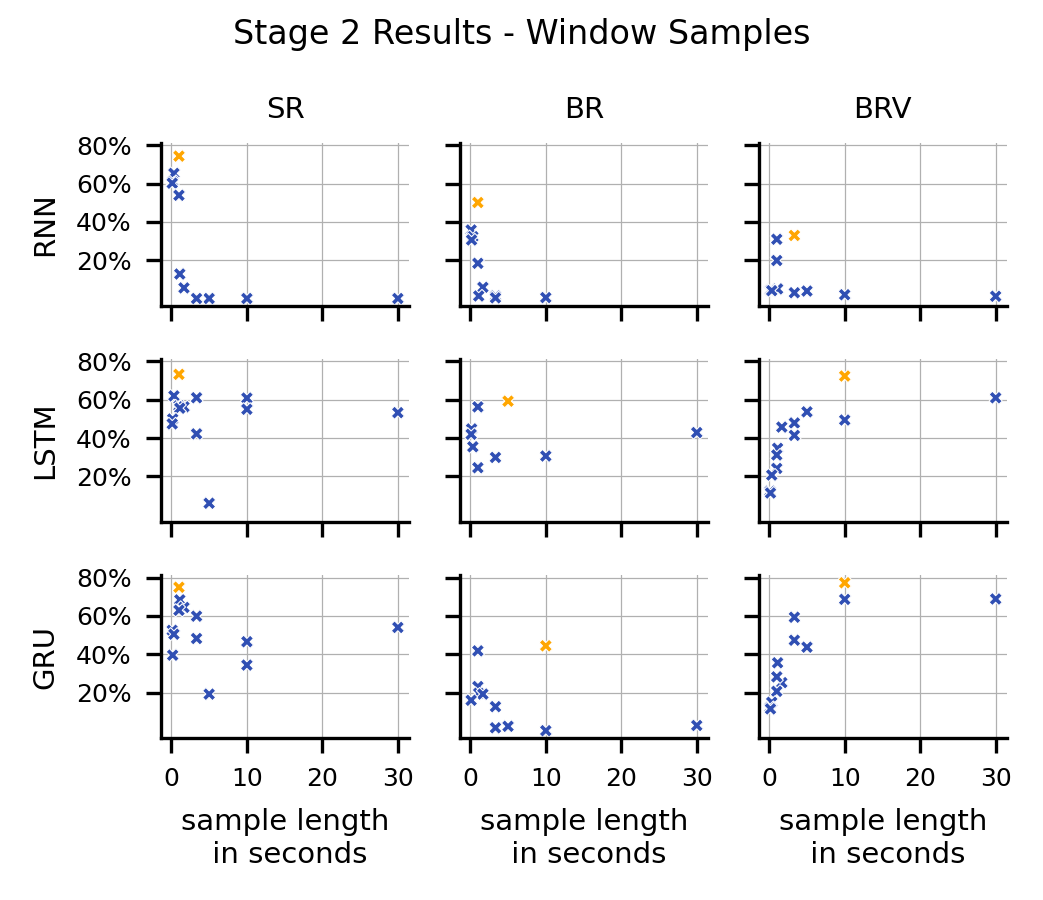}
		\caption{Results of the data hyperparameter search for windowed samples, $\textit{sample length} = \frac{\textit{window size}}{ \textit{fps}}$), indicating the achieved minimum accuracy of each configuration on the validation data in percent.}
		\label{fig:results_data_hps_windowed}
\end{figure}

\autoref{tab:configurations} lists the finally selected architectures and sample configurations after completion of both hyperparameter search stages.
During the first stage we retrained each of the nine architecture+data encoding combinations with at least 100 different hyperparameter configurations, totaling over 1700 training runs with a combined computation time of about 45 weeks.
In the second stage we retrained the configurations that achieved the highest minimum accuracy during the first stage with every sample configuration listed in \autoref{tab:data_hyperparameters}, the results are reported in \autoref{fig:results_data_hps_binned} and \autoref{fig:results_data_hps_windowed}.


\subsection{Evaluation}
\begin{table}
    \centering
    \caption{Mean and minimum accuracy of each model on the test data.}
    \label{tab:accuracy_report}
    \resizebox{\linewidth}{!}{
        \begin{tabular}{rrrr|rrrr}
\toprule
 & \multicolumn{3}{c}{mean accuracy} & \multicolumn{3}{c}{minimum accuracy} \\
 & SR (offset) & BR & BRV & SR (offset) & BR & BRV \\
\midrule
RF & 82\% (20\%) & 84\% & 53\% & 14\% (0\%) & 41\% & 11\% \\
MLP & 91\% ( 3\%) & 82\% & 62\% & 41\% (0\%) & 28\% & 22\% \\
FRNN & 90\% ( 5\%) & 82\% & 70\% & 42\% (0\%) & 12\% & 37\% \\
LSTM & 91\% ( 4\%) & 85\% & 82\% & 49\% (0\%) & 34\% & 40\% \\
GRU & 91\% ( 8\%) & 83\% & 86\% & 44\% (0\%) & 24\% & 56\% \\
\bottomrule
\end{tabular}

    }
\end{table}


The trained models are evaluated with the 34 test takes (one take for each subject with a minimum length of 5 minutes). We report the resulting accuracy scores for each model in \autoref{tab:accuracy_report}.

\begin{figure}[h]
	\input{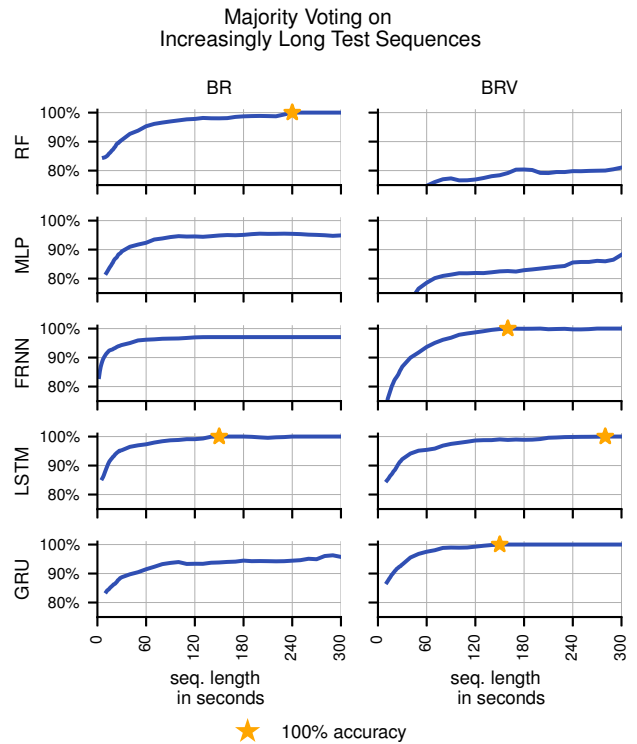}
	
    \caption{Clip classification accuracy on the test data for each model averaged over all subjects; e.g., the FRNN+BRV model achieves 90\% average classification accuracy of all 50 second clips, and classifies 100\% of all 160 second clips.}
	\label{fig:sequence_length_evaluation}
\end{figure}

	

High accuracies and little misclassifications suggest that the SR encoding generally outperforms the other data encodings.
However, it remains unclear if the models learned user-specific or session-specific characteristics.
As the hyperparameter search resulted in very short sample lengths for all SR models, we see reason to believe that the models are not learning actual movements, but rather memorize positions and orientations of the subjects.
To evaluate this, we created a second version of the SR test set, which we call \emph{SR offset}:
we shift each subject along both horizontal axes by adding 0.5 meters to the X- and Z- positions.
After re-evaluating the models on these SR offset data (reported as `offset' in \autoref{tab:accuracy_report}), the results show that SR models are completely thrown off, when the subjects are positioned differently than during training --- models trained on BR and BRV data remain unaffected, since the shifting has no effect on their features.

We assume that this constraint renders SR data less interesting for most scenarios and therefore focus on the BR and BRV encodings in the further analysis.
Here, the GRU+BRV model achieves with 86\% the highest mean accuracy on the test data.
All RNN models perform generally better than the RF or MLP models, with the exception of the RF+BR model that ranks third overall.

\autoref{fig:sequence_length_evaluation} reports the accuracies if we apply a majority approach to consider more than one sample for a prediction, as explained in section \ref{sec:prediction}.
Here, most models can identify all of the 34 subjects in the test takes correctly within 5 minutes, the only exception being the RF+SR, RNN+BR and MLP+BR models.

In general, most RNNs perform well: LSTM+BR, GRU+BRV, LSTM+BRV, FRNN+BRV all reach 100\% accuracy within 5 minutes.
The LSTM+BR and GRU+BRV require with 2.5 minutes the shortest period to reach 100\% accuracy.
Interestingly, the FRNN+BRV is not much worse and achieves 100\% accuracy in 160 seconds.
From the non-deep learning models only the RF+BR model achieves 100\% accuracy within 5 minutes (after 240 seconds).


\section{Discussion}

The results confirm that classification with SR, BR and BRV data sampled from arbitrary motion data is possible.
However, the SR encoding is problematic, since it includes a lot of straightforward session specific data (i.e., position and orientation of a user within the scene).
Models achieve high accuracy scores on the validation data that originate from the same session, but fail if users change location within the scene.
While this is arguably an expectable result, we think this is an issue worth highlighting:
if position or orientation of subjects is not guaranteed to match the training data during inference, for example when they can move outside the area the training data was recorded in, it becomes inevitable to train models on BR or BRV data.
That being said, some SR features, like the positional up-axis, remain interesting, since they can include important personal characteristics (e.g., body height) in XR settings where body proportions are not normalized. 

The machine learning models were able to also identify subjects with BR and BRV data.
The highest accuracies on the test data were reached with the BRV data.
The reason may be that models trained on BRV data are less prone to overfitting on predominant body postures of subjects in the training data.
Neural networks produced better results compared to random forest models on this dataset.
RF and RNNs work comparatively well on SR and BR data.
However, RNNs outperformed RF and MLP on the BRV encodings considerably.
We theorize that patterns in BRV encoded data are more difficult to learn and RNNs seem to profit from working directly on the raw sequence data.
In general, we believe that the BRV encoding provides a promising basis for robust user identification especially for multi-session scenarios and is an interesting topic for future work.

A majority voting approach on sequences with more than one sample improved the reliability of all models.
Here, BR and BRV data yield similar results.
The deep learning approaches performed generally well and profit from samples that span over a longer period of time.
We think that all three RNN architectures are viable models for the identification tasks, differences in the achieved accuracies might be eliminated with an even broader hyperparameter search.
Random Forest and MLP architectures were all outperformed by at least one RNN architecture, even though it has to be noted that the RF+BR model achieved considerable results.

The goal of this paper is to provide viable insights for future research in the field of user identification with motion data in XR contexts.
However, the used dataset originates from a full body motion capture system, and not from typical XR hardware --- so are our findings even valid for XR setups?
Since we specifically only selected features also available in XR setups, and other characteristics are comparable (e.g., frame rate, reference, etc.), we see only two potential objections: 
first, sensor-specific characteristics are different between a motion-capture system and a HTC Vive.
However, sensor-specific characteristics are also different between an HTC Vive and an Oculus Rift, so in either case models would have to be retrained on data from the respective hardware for optimal performance, and Miller R. et al. \cite{Miller2021} have shown that machine learning models can even work with multiple systems at the same time.
Second, user-specific characteristics could be more or less dominant in an XR scenario where the user wears HMD and controllers, compared to a full-body motion tracking scenario.
We think that this is much more an issue of the target application scenario than of the tracking hardware used.
There will be inherently much more movement, and therefore more potentially identifying behaviour, in more active settings where users need to walk around compared to more passive settings where users can sit down and may need to interact only occasionally.
Altogether, our findings about architecture and data encodings are not tied to the hardware it was recorded with and are therefore relevant for future research in the XR context.
In fact, the results presented can serve as a baseline for future research, which is an urgent desideratum in this field, since hardly any of the previous work published their data or code.



The normalization to a uniform skeleton of the movement data can currently be seen not as a limitation but as a benefit of our solution.
Since the models already solve the identification task with high accuracy on this --- in terms of the individual body proportions --- somewhat degraded dataset, it is safe to assume our solution would at least work as reliable on a dataset including information on body proportions that would originate from an actual XR setup.
Chances are, it would even show an improved performance.

Overall, we think the main limitation of this work is that there is only one session per user in this dataset.
Future work should therefore publish even larger datasets that contain at least two sessions per user, which allows to verify the findings of this work in a multi-session setting.



\section{Conclusion}

This article compared five machine learning architectures on three selections of movement data for user and avatar identification. Results on body-relative and velocity data promise to enable identification of subjects across multiple sessions. After a hyperparameter search LSTM+BR and GRU+BRV outperform competing models and can classify all subjects with an accuracy of 100\% within 150 seconds.

Altogether, our approach provides an effective foundation for behaviometric-based identification solutions, based on deep learning of arbitrary motion data sequences in XR contexts. We believe that such solutions will be increasingly important specifically for embodied encounters in social XR, where the look and appearance of avatars becomes more and more indistinguishable from the appearance of real human individuals.
In summary, this work a) compares three different data encodings, b) compares five different machine learning architectures, c) proposes a solution to find effective models and d) motivates efforts to create more expansive biometric datasets for behaviometric-based user identification.

\bibliographystyle{ieeetr}

\bibliography{references,marc,chris}
\end{document}